\documentclass{article}

\usepackage{arxiv}
\usepackage[utf8]{inputenc}
\usepackage[T1]{fontenc}
\usepackage{amsmath,amssymb}
\usepackage[numbers,sort&compress]{natbib}
\usepackage{authblk}
\usepackage{url}
\usepackage{hyperref}
\usepackage{microtype}
\usepackage{tabularx,booktabs,array,ragged2e}

\title{Why Linear Models Can Still Work in a Complex and Changing World}

\author[1,2]{Fangzhou Lin}
\author[2]{Dengzhe Hou}
\author[2]{Yinxing Li}
\author[2]{Kazunori D Yamada\thanks{Corresponding author: \texttt{yamada@tohoku.ac.jp}}}

\affil[1]{Texas A\&M University, USA}
\affil[2]{Tohoku University, Japan}

\date{October 2026}
\renewcommand{\headeright}{Preprint}
\renewcommand{\undertitle}{Preprint}
\renewcommand{\shorttitle}{Linear Models in a Complex and Changing World}

\hypersetup{
  pdftitle={Why Linear Models Can Still Work in a Complex and Changing World},
  pdfauthor={Fangzhou Lin, Dengzhe Hou, Yinxing Li, Kazunori D Yamada},
  pdfkeywords={linear models, partial observation, distribution shift, out-of-distribution prediction, best linear prediction, regularization, spurious correlation},
  colorlinks=true,
  linkcolor=blue,
  citecolor=blue,
  urlcolor=blue
}

\begin{document}
\maketitle

\begin{abstract}
Linear models can predict well in complex domains even when underlying mechanisms are nonlinear. We argue that this can occur because the complexity of the world exceeds the complexity that can be learned reliably from finite, partially observed data. Flexible models can exploit curvature, thresholds, and interactions that fit one hospital, period, population, policy setting, or measurement process but fail elsewhere. When this instability lies mainly in the nonlinear component, omitting it can give linear models lower prediction error in a new environment. We connect this claim to work on simple linear rules, bias and variance, regularization, best linear prediction, measurement error, distribution shift, invariance, Rashomon sets, and underspecification. Proposition 1 gives a simple squared-error case in which a nonlinear rule learned in the training environment raises target error because it transfers too weakly; finite-sample estimation error can strengthen this result. Partial observation alone is not enough: the link between observed variables and unobserved factors must vary across environments. Stable nonlinear structure favors flexible models, and linear models remain vulnerable to linear spurious relationships. Model complexity should therefore be chosen by cross-environment stability as well as training-data accuracy.

\end{abstract}


\section{Introduction}

Linear models continue to be used across complex domains. In behavioral judgment and clinical prediction, simple statistical or linear approaches have sometimes matched unaided expert judgment or more flexible alternatives in the studies reviewed \citep{dawes1989,grove2000,christodoulou2019}. Why should a linear model work when the underlying mechanisms are unlikely to consist of simple straight-line relationships? Flexible nonlinear models can represent linear relationships together with curvature, thresholds, and complex interactions. Expressive capacity alone therefore seems to favor flexibility. Interpretability and computational convenience are part of the answer, but this paper asks a predictive question: under what conditions can a linear model be less vulnerable than a flexible model when prediction moves to a new environment?

Answering this question requires distinguishing how complex the world is from whether the information needed to learn that complexity is present in the recorded data. When the relevant state is sufficiently observed and a nonlinear relationship remains stable across environments, a flexible model can learn it and outperform a linear model. Researchers, however, can rarely observe every factor that determines an outcome. Data contain measurement error, selection in who or what is observed, unrecorded factors, and features that occur only in a particular hospital or period. A more expressive model cannot recover information that was never recorded.

With finite data and partial observation, a flexible model can use detailed rules that appear only among the recorded variables in the training environment. A random split within the same hospital or period may show a predictive gain on held-out cases from that environment. Yet if the hospital, period, population, policy, measurement device, or recording practice changes, the observed variables may no longer carry the same information about the underlying state. The practical question is therefore not whether a nonlinear model class contains a better rule in principle. It is whether the rule selected from the available data remains useful after deployment conditions change.

Our central claim is deliberately conditional. The robustness of a linear model does not follow automatically from linearity. When data are finite, only part of the important state is observed, and complex rules found in the training data do not work sufficiently well in another environment, a linear model that does not use those rules may make fewer prediction errors in an unseen environment. The advantage does not come from poor source fit itself. It comes from restricting the fitted relationship in a way that excludes unstable curvature, thresholds, and interactions. Conversely, if a complex rule remains stable in another environment, there is no reason to discard it. A linear model is also not protected when the relationship that changes across environments is itself linear.

The components of this explanation are dispersed across research on improper linear models, clinical versus statistical prediction, bias and variance, regularization, best linear prediction, measurement error, environmental change, invariance, Rashomon sets, and underspecification. To our knowledge, they have not usually been organized around the single question of why a linear model can remain useful in a complex world that is only partly observed. Our first contribution is to synthesize these traditions and identify both the conditions favoring linear restriction and the cases in which the claim fails. Our second contribution is Proposition 1, which gives one simple setting in which the broader explanation can be established mathematically. It does not prove the paper's full claim for every possible setting. Finite data and partial observation are not themselves assumptions of Proposition 1; they are examined separately as reasons why unstable complex rules can arise.

In this paper, an environment means any condition that can change the characteristics of the data, including time, region, institution, population, policy, measurement device, or data-collection process. Prediction in an environment different from the training environment is called out-of-distribution (OOD) prediction. Throughout the paper, ``linear'' means linear relative to a fixed set of input features made from the recorded $X$. Changing that feature set changes what counts as linear.

\section{What Is Being Predicted When Important Factors Are Unobserved?}

We first explain the notation. Let $e$ index the period, institution, population, or other environment from which the data come. Let $Z$ denote the true state relevant to the outcome, only part of which is visible to the researcher. Let $X$ denote the variables that the researcher can actually record, and let $Y$ denote the outcome to be predicted. For example, $Z$ may be a patient's true health state, $X$ the available test results and medical records, and $Y$ a future symptom. Let $P_e$ denote the probability distribution that describes how data arise in environment $e$, and let $E_e$ denote an average, or expected value, under that distribution. Uppercase letters such as $X$ and $Y$ denote variables whose values have not yet been observed, whereas lowercase $x$ denotes an observed value.

These relationships can be written briefly as

\begin{equation*}
X=h_e(Z,\eta), \qquad Y=f(Z)+\varepsilon.
\end{equation*}

The first equation, $X=h_e(Z,\eta)$, means that the true state $Z$ passes through the environment-specific recording process $h_e$ and measurement noise $\eta$ to produce the observed value $X$. The subscript $e$ on $h_e$ allows the same underlying state to be recorded differently when, for example, the hospital or measurement device changes. The second equation, $Y=f(Z)+\varepsilon$, means that the outcome $Y$ is determined from the true state $Z$ through a rule $f$, together with an unpredictable outcome-side deviation $\varepsilon$. In this description, we assume $E_e[\varepsilon\mid X]=0$, so the deviation has mean zero after $X$ is observed.

A prediction model can use the observed $X$, but not the unobserved $Z$. We therefore write $m_e(x)$ for the average value of $Y$ among cases for which $X=x$ in environment $e$:

\begin{equation*}
m_e(x)=E_e[Y\mid X=x].
\end{equation*}

This expression means: among people or cases in environment $e$ whose observed value is $x$, what is the average outcome $Y$? It is called the conditional mean or regression function. When prediction performance is evaluated by average squared error, $m_e(x)$ is the theoretically best possible prediction. Thus, the object to be learned in an actual prediction problem is not the latent rule $f$ that generates the outcome from the unobserved $Z$. It is the function $m_e$ that predicts the outcome from the observed $X$.

Measurement error and unobserved factors make the shape of a complex relationship difficult to establish from data \citep{chen2011}. Theoretical work shows that partial observation can impose a predictability limit \citep{abeliuk2020}. In a large scientific mass collaboration, even many modeling teams achieved only limited accuracy for several life outcomes \citep{salganik2020}. Missing information does not, however, imply that $m_e$ is linear. The relationship after observation may remain nonlinear when, for example, the outcome changes sharply at a threshold, populations with different characteristics are mixed, or only selected individuals enter the data.

A mathematical example illustrates one special case. Its main conclusion is that, as the connection between the unobserved state and the observed value becomes weaker, complex higher-order patterns can weaken faster than a simple first-order pattern. This does not mean that a nonlinear relationship turns into a linear one.

For this paragraph only, let $Z$ and $X$ each be a scalar. Suppose that they have a joint Gaussian, or normal, distribution and are standardized to have mean zero and variance one. Let $r=\operatorname{Corr}(Z,X)$ be their correlation. Also suppose that the expected value of $f(Z)^2$ is finite. Using the probabilists' Hermite polynomials $H_k$, we can then separate $f$ into a constant component, a first-order component, a second-order component, and so on. The expansion holds in the $L^2$ sense, meaning in mean squared error, and can be written as

\begin{equation*}
f(Z)=\sum_{k\geq0}a_kH_k(Z).
\end{equation*}

Here, $a_k$ describes the size of the degree-$k$ component. For two jointly Gaussian variables, the following identities are known to hold \citep{severini2026}:

\begin{equation*}
E[H_k(Z)\mid X]=r^kH_k(X), \qquad
E[f(Z)\mid X]=\sum_{k\geq0}a_kr^kH_k(X).
\end{equation*}

Here, $E[\,\cdot\mid X]$ denotes an average after $X$ has been observed. On the right-hand side, the degree-$k$ component is multiplied by $r^k$. When $|r|<1$, its magnitude is multiplied by $|r|^k$, which decreases more rapidly as $k$ increases. Thus, if a first-order component was present originally, it can become relatively more prominent after observation.

This does not mean that adding noise creates a linear relationship. If the original relationship contains only a second-order component, that component becomes weaker but does not become first-order. The example shows only that, when an important state cannot be observed completely, complex components can become more difficult to distinguish than simpler ones. It does not show that the world becomes linear.

\section{What Previous Research Has Established}

The argument in this paper does not follow directly from any single body of research. Several research traditions support different parts of it (Table 1).

\begin{table}[htbp]
\centering
\caption{Literature synthesis and limits of inference}
\label{tab:literature-synthesis}
\small
\setlength{\tabcolsep}{4pt}
\renewcommand{\arraystretch}{1.25}
\begin{tabularx}{\textwidth}{@{}>{\RaggedRight\arraybackslash}X>{\RaggedRight\arraybackslash}X>{\RaggedRight\arraybackslash}X@{}}
\toprule
\textbf{Research area} & \textbf{What previous research establishes} & \textbf{What it does not establish by itself} \\
\midrule
Simple linear rules and mechanical prediction & Methods that do not estimate fine differences among weights can reduce estimation error and can sometimes be more accurate than human judgment or complex weighting & That linear models always perform better in unseen environments \\
\addlinespace[2pt]
Bias--variance and regularization & Restricting the relationships that can be learned can reduce the risk of learning accidental patterns from finite data & That models with many parameters are necessarily worse \\
\addlinespace[2pt]
Best linear prediction & The best linear approximation can be defined even when the true relationship is nonlinear & That the same linear approximation remains best across environments \\
\addlinespace[2pt]
Measurement error and unobserved factors & Missing information can make complex rules difficult to establish and can change the meaning of observed variables across environments & That missing information makes the prediction function linear \\
\addlinespace[2pt]
Environmental change and shortcut learning & A feature that is useful in the training environment can cease to work in the deployment environment & That every relationship that fails to transfer is nonlinear \\
\addlinespace[2pt]
Methods for learning relationships that remain stable across environments & Learning procedures can prepare for specified kinds of environmental change & That they can withstand every possible unseen change \\
\addlinespace[2pt]
Rashomon sets and underspecification & Models with similar training-environment accuracy can perform differently in another environment & That the simplest model is always best in an unseen environment \\
\bottomrule
\end{tabularx}
\end{table}

The first tradition concerns improper linear models. Rather than estimating a separate, finely tuned weight for each variable from the data, these methods use simple linear rules, such as assigning equal or unit weights to all variables. When data are limited and the estimation error in the weights is larger than the true differences in importance among the variables, a simple method using equal weights can outperform a method using estimated weights \citep{einhorn1975,dawes1979}. In other words, deliberately not learning small differences can sometimes reduce error. This literature alone, however, does not establish that linear models generally perform better in unseen environments.

The second tradition concerns the literature on clinical versus statistical, actuarial, or mechanical prediction. It compares expert judgment that combines information through experience and intuition with actuarial or mechanical prediction that applies the same prespecified equation, scoring rule, or table to every case. A representative review and meta-analysis found that predictions made by a fixed procedure often matched or exceeded the accuracy of expert judgment \citep{dawes1989,grove2000}. The important advantage here is the consistent reuse of the same procedure, not linearity itself. A nonlinear model can also make predictions by following a fixed procedure.

The third tradition concerns bias and variance. Because a flexible model can represent many shapes, finite data make it easier for the model to learn shapes that arose by chance \citep{geman1992}. Regularization reduces this problem by effectively restricting the shapes the model can learn. In some cases, regularization is mathematically equivalent to preparing for a specified form of data perturbation \citep{xu2009,bertsimas2018}. It does not follow, however, that a model is necessarily worse when it has more parameters. Models with extremely many parameters can sometimes predict well, as shown by work on double descent and benign overfitting \citep{belkin2019,bartlett2020}. What matters is not parameter count alone, but which relationships the learning procedure captures and which it suppresses.

The fourth tradition is best linear prediction. Even when the true relationship is nonlinear, one can define the linear prediction rule with the smallest mean squared error among all linear prediction rules. Suppose that there are $p$ observed variables, with values $x_1,\ldots,x_p$. A linear prediction based on these variables can be written as

\begin{equation*}
\beta_0+\beta_1x_1+\cdots+\beta_px_p.
\end{equation*}

The coefficients $\beta_0,\ldots,\beta_p$ determine how much each variable contributes to the prediction. Let $\mathcal L$ denote the collection of all prediction rules of this form obtained by varying these coefficients. The best linear predictor in environment $e$, denoted by $\ell_e^*$, is defined as

\begin{equation*}
\ell_e^*=\arg\min_{\ell\in\mathcal L}E_e[(Y-\ell(X))^2].
\end{equation*}

The notation $\arg\min$ means choosing the $\ell$ that makes the average error inside the brackets as small as possible. Mathematically, this operation selects the linear function closest to the true prediction function $m_e$ and is called an orthogonal projection \citep{white1980}. A linear model can therefore be used as the best linear approximation without assuming that the true relationship is linear. However, if the values taken by $X$, and their frequencies, change across environments, the coefficients of the best linear approximation can also change \citep{buja2019}. The existence of a best linear predictor and its invariance across environments are separate issues.

The fifth tradition concerns changes in the data between training and deployment. Let $s$ denote the training, or source, environment and $t$ the deployment, or target, environment. Covariate shift refers to the case in which the distribution of $X$ changes but the relationship between $X$ and $Y$ remains unchanged. In notation, $P_s^X\ne P_t^X$ but $P_s(Y\mid X)=P_t(Y\mid X)$ \citep{shimodaira2000}. If the range of $X$ that appears in the target environment is adequately represented in the training data and a suitable method is available, using a stable nonlinear relationship is not itself harmful.

The main concern of this paper is the case in which the relationship between $X$ and $Y$ itself changes. Flexible models can use shortcuts tied to backgrounds, institutions, measurement processes, or interactions among variables that are useful only in the training environment \citep{geirhos2020,sagawa2020,zhou2021}. Such shortcuts can also be useful in validation data randomly drawn from the training environment. Ordinary random splitting therefore cannot determine whether the learned rule will remain useful in another environment.

Methods that use differences among environments during training have also been studied. Invariant Causal Prediction searches for relationships that remain unchanged across multiple environments \citep{peters2016}. Anchor Regression balances fit to the training data against robustness to a prespecified type of environmental change \citep{rothenhausler2021}. Under particular mathematical conditions, a method with a linear estimation rule has been shown to achieve near-optimal worst-case performance even relative to a broader class that includes nonlinear methods \citep{lei2021}. Here, ``linear'' refers to the estimation rule in that study and is not identical to the class of linear prediction functions used throughout this paper. This is a result for specific conditions, not evidence that linear models always perform better. Research on heterogeneous covariate shift has likewise shown that resilience depends on both feature-specific shifts and the complexity of the function classes used for those features \citep{simchowitz2023}. Ultimately, a model restriction is useful only when it matches the environmental change that actually occurs.

Finally, many models can sometimes achieve nearly the same predictive accuracy. This collection is called a Rashomon set, and it can include simpler models \citep{breiman2001,semenova2022}. In addition, learning procedures with nearly identical performance on the same training data can produce substantially different results after the environment changes. This problem has been studied as underspecification \citep{damour2022}. Accuracy in the training environment alone may therefore be insufficient to determine which model has learned stable relationships.

\section{Under What Conditions Is a Linear Model Preferable?}

\subsection{An Ideal Case without Estimation Error}

Let $s$ denote the training, or source, environment and $t$ the deployment, or target, environment. To isolate one mechanism by which a linear model can be preferable, we deliberately impose simple conditions. The role of Proposition 1 is not to prove the broader claim synthesized from the preceding literature for every possible setting. It is to verify mathematically, in one explicit setting, that omitting a source-environment nonlinear rule can be beneficial when that rule does not transfer sufficiently to the target environment. Section 4.2 separately considers finite-data estimation error, and Section 4.3 considers how partial observation can generate unstable rules.

Assume that (i) the observed variables $X$ have the same probability distribution $P_X$ in the two environments, and (ii) $\mathcal L$ is the same collection of linear prediction rules in both environments. To ensure mathematically that a closest linear prediction rule exists, let $\mathcal L$ be a closed linear space. We also assume that every function used below has a finite expected squared value.

For two functions $u$ and $v$, define

\begin{equation*}
\langle u,v\rangle:=E_{P_X}[u(X)v(X)],\qquad
\lVert u\rVert^2:=\langle u,u\rangle.
\end{equation*}

The first quantity, $\langle u,v\rangle$, is the average product of $u(X)$ and $v(X)$ and describes how strongly the two functions point in the same direction overall. The second quantity, $\lVert u\rVert^2$, is the average of $u(X)^2$ and describes the overall size of the function $u$.

Next, assume that (iii), in each environment $e\in\{s,t\}$, the true prediction function $m_e$ can be decomposed as

\begin{equation*}
m_e(X)=\ell(X)+g_e(X).
\end{equation*}

Here, $\ell\in\mathcal L$ is a linear component shared by both environments. The function $g_e$ is the remaining nonlinear component in environment $e$; it may be the same or different across environments. To state precisely that no linear component remains in $g_e$, assume that $\langle g_e,q\rangle=0$ for every linear function $q\in\mathcal L$. This relation is called orthogonality.

Finally, assume that (iv) the observed outcome can be written as $Y=m_e(X)+\varepsilon_e$. The term $\varepsilon_e$ is an unpredictable random deviation whose average after $X$ is known is zero; that is, $E_e[\varepsilon_e\mid X]=0$.

Let $f_F=\ell+g_s$ be an ideal flexible predictor that uses the nonlinear component from the source environment, and let $f_L=\ell$ be an ideal linear predictor that uses only the shared linear component. The subscripts $F$ and $L$ stand for flexible and linear, respectively.

\textbf{Proposition 1 (comparison of prediction error in the target environment).} Let the mean squared error in the target environment be

\begin{equation*}
R_t(f)=E_t[(Y-f(X))^2].
\end{equation*}

This is the average, in target environment $t$, of the squared difference between the prediction $f(X)$ and the observed outcome $Y$. Under assumptions (i)--(iv), the difference between the errors of the flexible and linear predictors is

\begin{equation*}
R_t(f_F)-R_t(f_L)=\lVert g_s\rVert^2-2\langle g_s,g_t\rangle. \tag{1}
\end{equation*}

Consequently, if $g_s\ne0$ and

\begin{equation*}
A_{s,t}:=\frac{\langle g_s,g_t\rangle}{\lVert g_s\rVert^2}<\frac12,
\end{equation*}

the linear predictor has a smaller mean squared error in the target environment.

This result can be verified as follows. In the target environment, the component omitted by the linear predictor is $g_t$. The flexible predictor uses the source-environment rule $g_s$, so the remaining discrepancy is $g_t-g_s$. The prediction errors of the two models in the target environment can therefore be written as

\begin{equation*}
R_t(f_L)=\lVert g_t\rVert^2+E_t[\varepsilon_t^2],\qquad
R_t(f_F)=\lVert g_t-g_s\rVert^2+E_t[\varepsilon_t^2].
\end{equation*}

When the error of the linear predictor, $R_t(f_L)$, is subtracted from the error of the flexible predictor, $R_t(f_F)$, the common noise term cancels. Rearranging the remaining terms gives Equation (1).

The quantity $A_{s,t}$ is defined here to describe how strongly the nonlinear rule $g_s$ from the source environment continues to operate in the same direction as the nonlinear rule $g_t$ in the target environment. It is not an ordinary correlation coefficient or a proportion and therefore need not lie between zero and one. If $A_{s,t}=1$, the component of $g_t$ pointing in the same direction as $g_s$ has the same size as $g_s$, although $g_t$ may contain an additional component in another direction. If $A_{s,t}=0$, the target environment contains no rule pointing in the direction of $g_s$. If it is negative, the component of $g_t$ in the direction of $g_s$ points the other way.

What Proposition 1 directly establishes is that, under assumptions (i)--(iv), when this value is below $1/2$, the cost of carrying the source-environment nonlinear rule into the target environment exceeds the benefit of using it, and the linear predictor performs better. The threshold $1/2$ follows from the simple assumptions and error measure used here. It is not a universal boundary that applies to every prediction problem, nor does it define the broader claim of this paper.

Two cautions are necessary. First, we assumed that the distribution of $X$ is the same in the source and target environments. If the distribution of $X$ also changes, the equations must be modified so that the two models are compared under the same reference distribution. In that case, the best linear approximation in the source environment need not remain the best linear approximation in the target environment. Second, we assumed that the environment-dependent component $g_e$ contains no linear component. If the relationship that changes across environments is linear, the linear model is also affected.

\subsection{The Finite-Data Case}

Proposition 1 considered an ideal case in which the rule in the source environment can be learned without error. In practice, data are finite, so a fitted model also contains estimation error. Write the linear predictor learned from finite data as $\widehat f_L=\ell+\delta_L$, where $\delta_L\in\mathcal L$, and the flexible predictor as $\widehat f_F=\ell+g_s+\delta_F$. The hat indicates that the predictor has been estimated from finite data. The functions $\delta_L$ and $\delta_F$ are the errors produced by the respective learning procedures. Target risk is evaluated on new target cases that are independent of the sampled training data.

Let $E_{\mathrm{train}}$ denote an average over repeatedly drawing new training datasets and fitting the models again. To simplify the expression, make the following assumptions about how estimation errors align with the target-environment rules:

\begin{equation*}
E_{\mathrm{train}}\langle\delta_L,g_t\rangle=0,\qquad
E_{\mathrm{train}}\langle\delta_F,g_t-g_s\rangle=0.
\end{equation*}

These are strong simplifying assumptions, especially the condition on $\delta_F$; they are not automatic properties of flexible models. Without them, additional terms would remain in the expression below. Define

\begin{equation*}
V_L=E_{\mathrm{train}}\lVert\delta_L\rVert^2
\quad\text{and}\quad
V_F=E_{\mathrm{train}}\lVert\delta_F\rVert^2
\end{equation*}

as the average estimation errors of the linear and flexible models, respectively. Then

\begin{equation*}
E_{\mathrm{train}}[R_t(\widehat f_F)-R_t(\widehat f_L)]
=\lVert g_s\rVert^2-2\langle g_s,g_t\rangle+(V_F-V_L).
\end{equation*}

For $g_s\ne0$, the linear predictor has a smaller expected target mean squared error whenever

\begin{equation*}
A_{s,t}<\frac12+\frac{V_F-V_L}{2\lVert g_s\rVert^2}.
\end{equation*}

If $V_F>V_L$, the right-hand side is greater than $1/2$. This means that, as the flexible model's estimation error increases, the range of cases in which the linear model performs better expands, even when the nonlinear source rule remains somewhat useful in the target environment.

Limited data, many variables, large measurement error, and a large number of candidate interactions can increase the estimation error $V_F$ of a flexible model. Nevertheless, $V_F>V_L$ is not guaranteed. A sufficiently regularized flexible model can sometimes keep its estimation error small. Moreover, if the simplifying assumptions above do not hold, the additional error terms must be retained.

\subsection{Why Unobserved Factors Can Produce This Situation}

Let $U$ be an unobserved factor, and let $a$ be a scalar describing how strongly $U$ affects the outcome. Let $\ell_0(X)$ be a linear component that remains stable across environments. Suppose that the outcome is determined by

\begin{equation*}
Y=\ell_0(X)+aU+\varepsilon,\qquad E_e[\varepsilon\mid X]=0.
\end{equation*}

The learner cannot observe $U$ and must therefore infer it indirectly from $X$. If $E_e[U\mid X]$ denotes the average prediction of $U$ after $X$ is known, the prediction function for $Y$ based on $X$ is

\begin{equation*}
E_e[Y\mid X]=\ell_0(X)+aE_e[U\mid X].
\end{equation*}

Define

\begin{equation*}
q_e(X):=aE_e[U\mid X].
\end{equation*}

This function is an observed-data proxy: it is a cue constructed from the observed $X$ in place of the unobserved $U$. Let $\Pi_{\mathcal L}q_e$ denote the part of $q_e$ that can be represented linearly. The notation $\Pi_{\mathcal L}$ means selecting the linear function in $\mathcal L$ that is closest to a given function.

If this linear part is the same function $q_L$ in both environments, define

\begin{equation*}
\ell:=\ell_0+q_L,\qquad g_e:=q_e-q_L.
\end{equation*}

Then $\ell$ is the linear component shared across environments, while $g_e$ is the nonlinear remainder in environment $e$. This gives the decomposition $m_e=\ell+g_e$ used in Proposition 1.

When the hospital, policy, population, or measurement device changes, the effect $a$ of $U$ on the outcome may remain the same while $E_e[U\mid X]$ changes. Partial observation creates instability here only when this link between the observed $X$ and unobserved $U$ changes across environments. A complex use of $X$ that was informative in the source environment can then become weaker or operate in the opposite direction in another environment.

The presence of an unobserved factor does not, however, guarantee that a linear model is safe. If the linear part $\Pi_{\mathcal L}q_e$ also changes across environments, the linear model can learn an unstable relationship as well. Unobserved factors provide one mechanism that can create the situation examined in this paper; by themselves, they neither prove that linearity is correct nor provide a sufficient reason to fit a linear model.

\section{What Follows from This Analysis}

A linear model summarizes what can be predicted from the observed $X$ while restricting the relationship to a linear form. If the detailed rules that fail in other environments lie mainly in the nonlinear part, the inability to represent them can protect prediction. Conversely, if the discarded nonlinear rules are stable across environments, the same restriction merely causes underfitting.

This analysis also changes how validation results should be interpreted. If a flexible model performs better under a random split, it may genuinely be better for new data drawn from the same environment. That result alone, however, does not show that it will remain better when the time period, institution, population, policy, or measurement method changes. Validation conducted entirely within the training environment measures performance in that environment. It does not measure stability across environments. When several models have nearly the same accuracy, they may rely on different rules \citep{breiman2001,semenova2022}. Models that are equivalent in the training environment may then perform differently in the environment where they are used \citep{damour2022}. When there is little evidence that complex rules will continue to work elsewhere, choosing a simpler model can therefore have a predictive justification that is separate from ease of explanation.

The choice is not limited to either a strictly linear model or a highly complex model. Intermediate levels of complexity are available, including generalized additive models, spline models that represent smooth curves, models that use only a small number of interactions, and shallow decision trees. These models can retain only the curvature that the data support and exclude complexity for which the evidence is weak. A linear model with a very large number of variables may also be unstable, while a strongly regularized nonlinear model may be simpler in practice. The comparison should therefore not be based only on the labels "linear" and "nonlinear." It should also consider the variables actually used, regularization, tuning procedures, and the environments used for validation. With sufficient data from multiple environments, a flexible model may preserve stable nonlinear relationships while suppressing unstable cues. A strictly linear model is not the only way to construct a robust predictor.

\section{When the Claim Does Not Hold}

First, nonlinear relationships should be used when they are stable across environments. For example, if $E[Y\mid X,e]=\sin(X)$ holds in every environment and enough data are available, an appropriate nonlinear model will be more accurate than a linear model. Discarding this curvature is not a way to improve robustness; it is simply underfitting. An interpretable model also need not be strictly linear. Generalized additive models and models with a limited number of interactions can represent relationships in a form that people can understand \citep{caruana2015,rudin2019}.

Second, a spurious relationship need not be nonlinear. If an environment-specific cue is linearly associated with $Y$, a linear model can learn it as well.

Third, even when the relationship between $X$ and $Y$ does not change, the best linear approximation may change if the frequency or range of $X$ changes. When the true relationship is nonlinear, the slope of the approximating line depends on which parts of the range receive the most weight.

Fourth, a properly regularized flexible model can select an almost linear relationship when appropriate. With multiple training environments, sufficient data, and a learning method designed for the kinds of environmental change that are likely to occur, a flexible model can suppress unstable rules while retaining stable nonlinear relationships. The relevant comparison is not with an ideal nonlinear model that somehow knows everything. It is with the flexible model that is actually selected from the available training data. A review of clinical prediction studies found that machine-learning methods did not consistently outperform logistic regression \citep{christodoulou2019}. This is not a law stating that linear models are always superior. It is evidence that comparisons must be fair and carefully designed.

Fifth, any claim of robustness must specify the environmental changes against which robustness is required. Anchor Regression, for example, protects against a prespecified type of change; it does not protect against every unexpected change. No model is always best in every possible deployment environment.

Finally, Proposition 1 concerns the prediction of numerical outcomes under mean squared error. Classification, rare but severe failures, the accuracy of predicted probabilities, and settings in which overprediction and underprediction have different costs require different analyses. The argument in this paper applies only when the part excluded by the model overlaps with the part that is unstable across environments.

\section{What Empirical Studies Should Examine}

When selecting a model, researchers should compare models whose complexity increases gradually. One possible sequence is a model using only the mean, a linear model, a generalized additive model, a model with a small number of interactions, tree or kernel methods, and then more flexible learners. Researchers should examine not only the accuracy score, but also which curve, threshold, or interaction produced the improvement. They should then determine whether that improvement remains in other environments. Nearly tied models require particular attention because they may rely on different rules.

The validation design should match the conditions that are expected to change at deployment. If a model will be used in a future period, another hospital, another region, another population, or under a different policy or device, each such environment should be held out as a whole for evaluation. Randomly splitting individual records is not sufficient. Reports should include not only average performance, but also error in each environment, error in the worst-performing environment, agreement between predicted probabilities and observed frequencies, whether the ranking of models changes across environments, and the uncertainty in performance differences.

When information identifying the environments is unavailable, researchers can examine how much the results change across repeated samples, how much coefficients or estimated variable effects vary, and how accuracy changes as the sample size increases. These analyses may reveal warning signs, but they cannot replace evaluation in genuinely different environments.

The process by which the data were produced should also be examined. Which recorded variables serve as substitutes for unobserved states? Would the meaning of those variables change if the hospital, institution, device, or person responsible for recording the data changed? In some cases, improving the measurement process or collecting an important additional variable may be more effective than choosing a different model class. When data from multiple environments are available, methods that emphasize relationships shared across environments, Group DRO for reducing the worst group-specific error, and Anchor Regression for preparing for specified environmental changes should also be compared with a simple linear model. The aim is not to discard every nonlinear relationship. It is to use nonlinear relationships that have been shown to remain stable in other environments.

Empirical studies should report not only how much additional complexity improves accuracy in the training environment, but also how much of that improvement remains in other environments. They should not attribute every failure in another environment simply to model complexity. The gain from using a complex relationship must be distinguished from the error introduced by learning that relationship from finite data. The environmental changes used for evaluation should reflect actual changes in institutions, policies, populations, or measurement methods. Merely adding artificial noise to the data may test a different kind of robustness from the one that matters in practice.

Stronger evidence requires prospective comparison in a new environment whose outcomes have not yet been observed. The linear and flexible models should be fixed before the outcomes in that environment are examined. Differences between the new and training environments in measurement methods and in which people enter the data should be recorded. Researchers should also state in advance which additional rules used by the flexible model they expect to remain valid. The sample size should be large enough to estimate the range of error within each environment, not only to support a comparison after pooling all environments. When outcomes take a long time to become available, changes in the range of input variables, discrepancies in predicted probabilities, and changes in estimated variable effects can be monitored in the meantime. Such checks, however, cannot replace validation using the final outcomes.

The consequences for actual decisions should also be compared. A small difference in mean squared error may have little practical importance. By contrast, unstable predictions near a threshold for treatment, lending, or resource allocation may change the decision itself. Model complexity should therefore be evaluated in relation to the loss that matters in practice, the consequences for different groups, and the need to recalibrate the model after deployment. It is also informative when the additional contribution of nonlinear relationships repeatedly remains across several realistic environments. In that case, a strictly linear restriction causes an unnecessary loss of accuracy.

This research agenda can be summarized by three questions. First, which rules continue to operate in the same way when the environment changes? Second, which environmental changes have not yet been observed but are realistically possible? Third, how much evidence is needed before a complex rule should be trusted? The appropriate answer may be a linear model, a model with limited curvature, or a highly flexible model. Complexity should be selected by degree, rather than treated as a binary choice between linear and nonlinear models.

\section{Discussion}

Useful performance by a linear model in a complex domain does not imply that the mechanisms of the world are linear. The complexity of the world and the complexity that can be learned reliably from recorded data are different. A narrower predictive justification is available: a linear model can exclude detailed rules that the available data cannot establish as stable in other environments. Restricted source fit helps only when the lost fit comes from discarding unstable complexity; poorer fit is not intrinsically valuable.

Research on simple linear rules has shown that there are settings in which it is useful not to learn fine differences among weights. Research on the bias--variance tradeoff explains why learning complex rules from finite data can increase error. Work on best linear prediction shows that a best linear approximation can be defined even when the underlying relationship is nonlinear. Research on measurement error shows why the underlying mechanism must be distinguished from the relationship that can be predicted from observed variables. Research on environmental change, invariance, the Rashomon set, and underspecification shows that performing well in the training environment is not the same as performing well in the environment where the model will be used.

Taken together, the literature supports the broader claim that a linear model can be advantageous when the rules that vary across environments lie mainly in the nonlinear part, while the linear part is comparatively stable. Suppose further that data from the training environment alone cannot identify which nonlinear rules are unstable. Proposition 1 does not prove this entire claim. It uses a measure of how similarly the nonlinear rule learned in the training environment operates in the deployment environment to establish one case, under simple conditions, in which the linear model has lower error. The threshold $1/2$ in Proposition 1 follows only from that setting. With finite data, the error involved in estimating the complex rule must be considered separately. Partial observation is one mechanism that can change the predictive meaning of observed variables across environments and thereby generate unstable rules.

This result does not claim that higher-order relationships are generally unstable. Nonlinear laws in physics or biology may remain stable across environments. Conversely, a linear relationship may change when a policy changes or when the process determining who enters the data changes.

A linear model should therefore be treated neither as a universally correct baseline nor as a claim that reality is linear. It is one option on a continuum of model complexity. When the likely environmental changes are known and sufficient data from multiple environments are available, a flexible model may make use of a larger amount of stable structure. When little is known about environmental change, a linear model may be a cautious choice. Even then, researchers must examine whether its linear relationships are themselves stable and whether the data-recording process creates problems. The most important question is not whether reality is linear. It is how much of the complex structure found in the observed data will remain in the environments where the predictions are actually used.

\section{Conclusion}

Linear models sometimes achieve useful predictive performance in complex domains. This paper locates the explanation not in a simple world, but in the difference between the complexity of the world and the complexity that recorded data can support reliably. When data are finite, only some important factors are observed, and the link between observed and unobserved factors changes across environments, complex rules found in the training data may not work well enough elsewhere. If this instability lies mainly in the nonlinear component, a linear model that does not use those rules may make fewer prediction errors in an unfamiliar environment. The advantage does not arise from poorer training fit itself. It arises when linear restriction prevents a model from carrying nontransferable complexity into the target environment. This is the conditional explanation obtained by synthesizing the research reviewed in this paper. Proposition 1 does not prove the explanation for every possible setting; it provides one simple setting in which it holds. The numerical threshold obtained there is specific to that setting. If nonlinear rules are stable across environments, however, a flexible model should be used. If the relationships that change across environments are linear, a linear model offers no protection. The affirmative reason for using a linear model is that limited observations may not yet justify carrying complex rules into environments in which their validity has not been established.

\renewcommand{\refname}{References}
\bibliographystyle{unsrtnat}
\bibliography{references}

\end{document}